# Quantum-Grassmann-Plücker Token Mixing for Deep Learning-Based Post-Disaster Damage Assessment

Kooroush Farahkhah[1], Umut Lagap[2], Taha Rezaei[3], Saman Ghaffarian[4]

[1] Independent Researcher, Tabriz, Iran.

[2] Centre for Urban Resilience and Analytics (CURA), Georgia Institute of Technology (Georgia Tech), US.

[3] Department of Civil and Environmental Engineering, University of Exeter, Exeter, UK.

[4] Department of Risk and Disaster Reduction (RDR), University College London (UCL), UK.

## Abstract

Timely post-disaster building damage assessment from satellite imagery is a critical engineering decision support task, yet it remains constrained by class imbalance, ambiguous intermediate damage states, and limited cross-event transferability. This study presents, to our knowledge, the first application of Grassmann–Plücker (GP) token mixing to computer vision and introduces two extensions for image classification: the Quantum-inspired Grassmann–Plücker (QGP) head and the Hybrid Quantum Machine Learning Grassmann–Plücker (HQML-GP) head. The GP head represents multiscale relationships among image patch tokens by encoding subspaces formed by token pairs with Plücker coordinates; QGP enriches these coordinates with amplitude-derived probability features, whereas HQML-GP incorporates expectation values generated by a simulated quantum circuit into the geometric token representation. Paired pre- and post-event image patches from the public xBD tornado dataset were processed using a frozen six-channel Vision Transformer base encoder. The three proposed heads were compared with multilayer perceptron and Transformer baselines under identical training, checkpoint selection, and evaluation protocols. Joplin and Moore tornado samples were used for model development and seen-event testing, while Tuscaloosa was reserved for unseen-event evaluation.

QGP led both test sets in accuracy and macro-F1: 83.46% and 64.50% for the seen events, and 66.45% and 52.70% for the unseen event. Although HQML-GP obtained the highest validation macro-F1 of 65.63%, it did not surpass QGP on either test set and required substantially longer training. These results establish GP token mixing as a competitive attention-free alternative to conventional Transformer-based token mixing for paired satellite image damage classification.

# 1. Introduction

Natural hazard-induced disasters create an urgent need for accurate and timely information on building damage to minimize the adverse effects of the events [1,2]. Although field surveys remain indispensable, they are often slow, costly, hazardous, and spatially incomplete during the immediate aftermath of an event [3]. Remote sensing provides a complementary source of evidence because pre- and post-disaster satellite imagery can cover large affected areas and support rapid assessment by emergency managers, humanitarian organisations, insurers, and recovery planners [4,5]. For tornado and severe-wind events, this need is particularly pressing since damage is often spatially discontinuous, highly localised, and influenced by building vulnerability, exposure, and local wind conditions [6–8]. Automated building-level damage classification is therefore not merely a computer vision task, but a decision-support problem with direct implications for response prioritisation, recovery planning, and disaster risk reduction [9,10].

Deep learning has become a prominent approach to automated post-disaster damage assessment because convolutional, attention-based, and transformer-based architectures can learn complex visual representations from very-high-resolution imagery [11,12]. The xBD dataset has enabled systematic benchmarking by providing paired pre- and post-event satellite images, building polygons, ordinal damage labels, and metadata across multiple disaster types [13,14]. However, building damage classification continues to face three persistent challenges: severe damage classes are relatively scarce [15], intermediate damage states are visually ambiguous [16], and models trained on one disaster event may generalise poorly to another [17]. In addition, fully retraining large visual backbones for each new disaster may be computationally demanding and impractical when event-specific labels are limited [18]. These challenges motivate parameter-efficient approaches that can adapt pretrained models while seeking to improve balanced performance across classes and cross-event transferability [19].

Vision Transformers provide a natural representation framework for paired building damage assessment by representing images as sequences of patch tokens. In a pre- and post-event setting, these tokens can encode roof conditions, debris patterns, surrounding damage, structural alterations, and other change-related cues [20,21]. However, damage evidence is often relational rather than isolated. A displaced roof edge, partial collapse, or debris field may only become meaningful when interpreted in relation to neighbouring image regions and the corresponding pre-event condition [22]. Conventional mean pooling does not explicitly preserve such relationships [23], while self-attention learns token dependencies through dense attention weights without explicitly representing their geometric subspace structure. This motivates geometry-aware token mixing, in which relationships between patch tokens are represented as low dimensional subspaces [24]. Chong [25] introduced this mechanism for language modelling and natural language inference, where it achieved competitive performance without explicit attention. Nevertheless, that work was limited to text sequences; it did not apply GP mixing to visual patch tokens or image classification.

This study moves Grassmann–Plücker token mixing from language to computer vision, particularly paired satellite image damage assessment and extends it in two directions designed here. A frozen six-channel ViT-B/16 encoder extracts visual tokens from concatenated pre- and post-event image patches, while trainable classification heads determine how these tokens are mixed and aggregated for damage classification [26–28]. Freezing the encoder serves two purposes. First, it reduces the number of trainable parameters and supports rapid adaptation when event-specific data are limited [20]. Second, it isolates the contribution of the classification head architecture, enabling a controlled comparison between conventional pooling, attention-based token mixing, geometry-aware token mixing, and its quantum-inspired and hybrid extensions. The objective is not to replace fully trainable vision architectures, but to determine whether structured head-level adaptation can provide an efficient and transferable alternative for post-disaster damage assessment. Quantum-inspired and hybrid quantum-classical mechanisms are examined as optional feature enrichment strategies within the Grassmann–Plücker framework. Quantum machine learning has demonstrated that structured classical data can be embedded into high-dimensional feature spaces and transformed through nonlinear operations or parameterised quantum circuits [29,30]. This concept is compatible with Grassmann–Plücker representations because Plücker coordinates already provide compact geometric descriptors of token pair subspaces [31]. Rather than applying quantum processing to the complete vision architecture, the present study restricts quantum-inspired and hybrid operations to the classification head. This design preserves the frozen backbone comparison and allows the additional value of each enrichment mechanism to be evaluated directly. Two extensions of the standard Grassmann–Plücker head are evaluated. The Quantum-inspired Grassmann–Plücker head remains entirely classical and enriches the Plücker coordinate representation using amplitude-derived probability features. The Hybrid Quantum Machine Learning Grassmann–Plücker head additionally processes a global Plücker descriptor through a small differentiable quantum circuit and injects the resulting expectation values into the geometric token representation.

The central research questions are whether GP token mixing provides a competitive attention-free design for post-disaster building damage classification, and whether the developed QGP and HQML-GP extensions add value to this geometric representation. Building on standard Grassmann–Plücker token mixing as a geometric foundation, the study examines whether amplitude-derived probability features and expectation values obtained from a simulated quantum circuit improve predictive performance and cross-event generalisation beyond standard GP, mean pooling, and Transformer attention baselines. The evaluation focuses on tornado events in xBD to reduce hazard heterogeneity. Models are tested both on events represented during training and on a held-out tornado event, enabling assessment of within-event performance, cross-event transferability, class-specific behaviour, and computational efficiency. The main contributions of this paper are as follows:

1. A Grassmann–Plücker token mixing head is adapted to paired pre- and post-event satellite image damage classification using a frozen six-channel ViT-B/16 encoder.
2. The study introduces a geometry-aware head-level adaptation strategy that explicitly represents multi-scale relationships between patch tokens through Plücker coordinate subspace descriptors.
3. A classical Quantum-inspired Grassmann–Plücker extension is developed using amplitude-derived probability features.
4. A hybrid quantum-classical Grassmann–Plücker extension is developed by combining global Plücker descriptors with differentiable quantum circuit expectation values.
5. All proposed heads are compared with MLP and Transformer baselines under identical backbone, data-split, optimisation, checkpoint selection, and evaluation conditions.

6. The study evaluates both seen- and unseen-event performance and reports predictive accuracy, class-specific behaviour, trainable parameters, training cost, and inference speed.

## 2. Background

Deep learning has become an important analytical approach in disaster risk management, enabling the rapid extraction of spatial, temporal, and semantic information from large volumes of remote sensing imagery, videos, sensor measurements, and other heterogeneous data [32]. Previous studies have applied deep learning to wildfire prediction, fire detection, flood risk mapping, damage classification, and recovery monitoring [33]. Zhong et al. [34], for example, developed a deep learning-based digital twin framework for wildfire prediction, while Talaat and ZainEldin [35] applied YOLOv8 to urban fire detection. Chaudhuri and Bose [36] used convolutional neural networks to detect potential survivors in earthquake debris imagery, and Ahmed et al. [37] applied DeepLabv3 to satellite-based flood risk evaluation. In the specific context of building damage assessment, Valentijn et al. demonstrated the capability of CNNs to classify damage severity from high-resolution satellite imagery across multiple disaster types [38]. However, many existing applications rely on specialised end-to-end architectures that may require substantial retraining when transferred to new disaster conditions [39,40].

Very-high-resolution optical satellite imagery is particularly valuable for post-disaster building damage assessment as it enables building-scale analysis over large affected areas [41]. Earlier remote sensing approaches relied on visual interpretation, GIS-based analysis, object-based image analysis, and conventional change detection techniques [2]. Although these methods remain useful for expert interpretation and contextual validation, they are difficult to scale when rapid building-level information is required across extensive disaster footprints [9]. The release of the xBD dataset and the xView2 challenge substantially advanced the field by providing paired pre- and post-disaster satellite imagery, building footprints, and ordinal damage labels across multiple disaster events [13,14]. The dataset enabled systematic benchmarking and encouraged the development of models that jointly address building localisation and damage classification [40]. In the meantime, xBD highlighted several persistent challenges, including severe class imbalance, visual ambiguity between intermediate damage states, label uncertainty, and limited generalisation across disaster events [14,41,42]. These challenges are especially relevant to tornado and severe wind damage, which is often highly localised, spatially discontinuous, and influenced by building characteristics, surrounding debris, local exposure, and image acquisition conditions [43,44].

A major body of research has sought to improve post-disaster building damage assessment through CNN-based, Siamese, U-Net-based, and multistage architectures. Weber and Kané [45] formulated building damage assessment as a semantic segmentation task, demonstrating the value of dense prediction for spatially explicit damage mapping. Gupta and Shah [46] proposed RescueNet, a joint building segmentation and damage assessment model that combines localisation-aware learning with a multiheaded architecture. Zhang et al. [39] introduced LRBNet, which integrates Siamese feature extraction, U-Net++, lightweight compression, and channel attention for paired image analysis. Deng and Wang [40] further enhanced U-Net-style damage assessment models using additional skip connections, asymmetric convolution blocks, and shuffle attention. These studies reflect the importance of comparing pre- and post-event imagery rather than analysing each image independently [47]. However, their improvements are primarily achieved through increasingly specialised feature extraction and fusion architectures, which may increase computational requirements and complicate adaptation to new disaster events.

Attention mechanisms have also been widely adopted in post-disaster damage assessment because they can emphasise informative visual regions while suppressing irrelevant background variation [32]. Shen et al. [48] developed BDANet using multiscale feature fusion and cross-directional attention to improve building localisation and damage classification. Xia et al. [16] subsequently applied a two-stage BDANet-style framework to ultra-high-resolution imagery from the 2023 Turkey–Syria earthquakes, combining pre-event building extraction with pre/post-event feature fusion. Oludare et al. [49] proposed ATS-HRNet by integrating high-resolution network components with criss-cross attention, while Braik and Koliou [9] combined CNN-based classification with GIS analysis to preserve geospatial context. These studies indicate that attention and contextual modelling can improve damage assessment. Nevertheless, additional attention modules may increase model complexity, and strong aggregate accuracy does not necessarily imply reliable recognition of rare damage classes or robust transfer to unseen events [38,41].

Recent reliability-oriented research has therefore placed greater emphasis on class-specific performance, model trustworthiness, and behaviour under class imbalance. Lagap et al. [41] systematically evaluated channel, spatial, and multi-head attention mechanisms for post-disaster damage assessment using xBD tornado cases. Their results showed that multi-head attention could improve recognition of high-severity damage, whereas some channel- and spatial-attention configurations produced less reliable feature emphasis. In addition, their study demonstrated that attention mechanisms are not automatically beneficial and should be assessed using both aggregate and class-specific metrics. Since post-disaster datasets are commonly imbalanced and intermediate damage categories are visually ambiguous, accuracy alone may conceal poor performance on operationally important minority classes [14]. Metrics such as macro-F1, per-class F1, balanced accuracy, and MCC are therefore important for evaluating the reliability of damage classification systems [50]. This consideration directly motivates the class-specific and seen/unseen-event evaluation adopted in the present study [51].

Transformer-based architectures represent another important development in post-disaster damage assessment. The original Transformer introduced self-attention as a general mechanism for modelling relationships among sequential elements [52], while the Vision Transformer reformulated images as sequences of patch tokens [53]. In remote sensing, transformer-based change detection models such as BIT [54] and ChangeFormer [55] have demonstrated the ability to capture broad spatial context and temporal relationships between image pairs. For building damage assessment, Kaur et al. [20] proposed a hierarchical transformer architecture that combines multiscale spatial features with temporal differences between pre- and post-disaster imagery. DamageCAT, introduced by Xiao and Mostafavi, further illustrates the growing role of transformer-based frameworks in structured post-disaster damage categorisation [12]. Although transformers can model long range relationships between image regions, fully fine tuning large transformer backbones may be computationally demanding when labelled event-specific data are limited. This limitation makes frozen-backbone and head-level adaptation strategies particularly relevant to rapid post-disaster deployment [56].

The digital twin-enabled recovery monitoring framework proposed by Lagap and Ghaffarian [33] further highlights the importance of transferable, efficient, and adaptable learning systems for disaster applications. Their work combines transfer learning, attention mechanisms, and explainable AI to support the monitoring of changing post-disaster conditions. Although the present study focuses on initial building damage classification rather than longitudinal recovery monitoring, the broader requirement for efficient model adaptation remains relevant. A frozen visual backbone can preserve a stable pretrained representation, while a smaller task-specific head can be trained or replaced for a new disaster assessment task [57]. This approach avoids retraining the complete feature extractor and enables the contribution of the classification head to be examined independently. Head-level adaptation is therefore not only an implementation choice in this study, but also a controlled strategy for assessing how different token mixing mechanisms use a shared visual representation [58].

Despite the progress achieved through convolution, feature fusion, attention, and transformers, most post-disaster damage assessment models rely on conventional pooling, self-attention, or dense fusion operations to aggregate visual information [39]. These mechanisms are effective, but they do not explicitly represent the geometric subspaces spanned by pairs of local patch token embeddings. Damage evidence is frequently relational: a displaced roof edge, debris field, partial collapse, or shadow pattern may only be meaningful when interpreted relative to adjacent image regions and the corresponding pre-event condition [16]. Grassmann manifolds represent linear subspaces rather than individual feature vectors and have previously been used in imageset recognition, subspace learning, and manifold-aware neural networks [59,60]. Grassmann–Plücker token mixing extends this concept to tokenised representations by mapping token pairs to low-dimensional subspaces and encoding them through Plücker coordinates [25]. Unlike conventional token pooling, this formulation explicitly represents pairwise token relationships. It is therefore well suited to paired building damage assessment, where damage patterns often emerge from relationships among neighbouring and multi-scale image regions rather than from isolated patch appearance.

Quantum-inspired and hybrid quantum-classical machine learning provide complementary mechanisms for enriching structured feature representations. Quantum machine learning investigates how classical data can be embedded into high-dimensional Hilbert spaces and transformed through parameterised quantum circuits or related nonlinear feature mappings [29,30,61]. Hybrid quantum-classical approaches incorporate small quantum circuits into otherwise classical learning pipelines [62], whereas quantum-inspired approaches remain entirely classical but adopt concepts such as amplitude normalisation and probability-based representations [63]. These mechanisms are conceptually compatible with Grassmann–Plücker token mixing because Plücker coordinates already provide compact geometric descriptors of token pair subspaces [31]. In the present study, quantum-inspired probability features and simulated quantum circuit expectation values are therefore evaluated as extensions of the standard GP representation. Their purpose is to determine whether additional nonlinear feature enrichment provides measurable benefits beyond geometry-aware token mixing, rather than to demonstrate quantum computational advantage.

The existing literature has substantially advanced post-disaster building damage assessment through CNNs, Siamese networks, U-Net variants, attention mechanisms, hierarchical transformers, and multi-temporal feature-fusion strategies [16,20,39,48]. However, many studies focus on modifying or fine-tuning the feature-extraction backbone, while classification heads commonly rely on mean pooling, attention, or conventional feed-forward aggregation [12,32,41]. Comparatively less attention has been given to head-level adaptation mechanisms that explicitly represent relationships between patch tokens through geometric subspaces. Grassmann–Plücker token mixing has not been systematically evaluated for paired pre- and post-event building damage classification under a frozen-backbone protocol. It also remains unclear whether classical quantum-inspired enrichment or hybrid quantum circuit derived features provide additional value beyond the standard geometric representation. To address these gaps, the present study develops, adopts, and evaluates GP, QGP, and HQML-GP heads on top of a common frozen six-channel ViT-B/16 encoder and compares them with conventional MLP and Transformer heads under identical training, model-selection, and seen/unseen-event evaluation conditions (Figure 1).

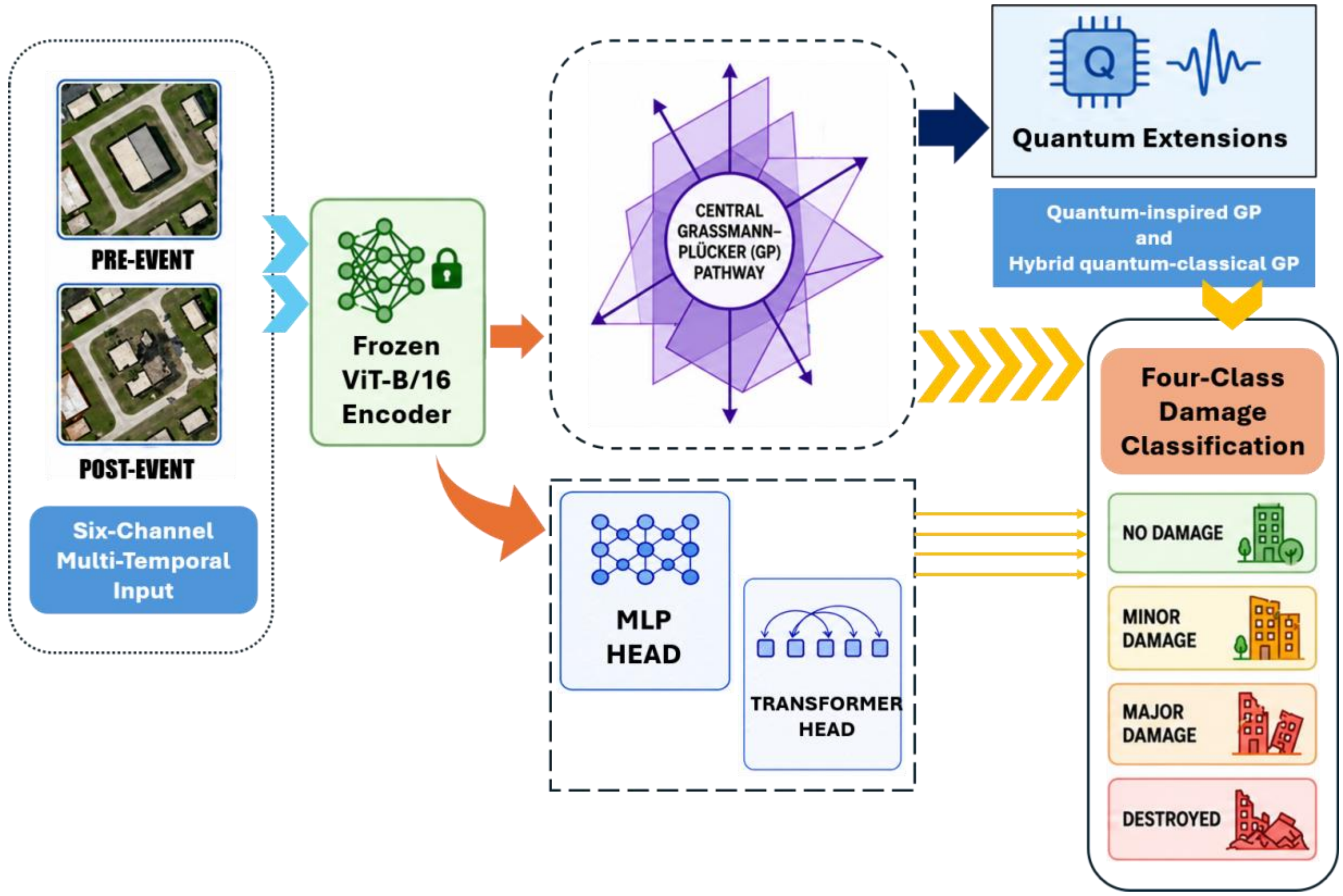

***Figure 1***. *Overview of the frozen-backbone head-level adaptation framework for post-disaster building damage classification. Paired pre-event and post-event image patches are concatenated into a six-channel input and processed by a common frozen ViT-B/16 encoder. The resulting patch tokens are supplied independently to MLP, Transformer, GP, QGP, and HQML-GP classification heads. The models predict four damage classes and are evaluated on seen-event and unseen-event test sets.*

# 3. Materials and Methods

## 3.1. Study design

This study evaluates whether geometry-aware, quantum-inspired, and hybrid quantum-classical token mixing heads can improve post-disaster building damage classification within a frozen-backbone adaptation framework. Each sample consists of paired pre-event and post-event image patches centred on an individual building. The two RGB images are concatenated along the channel dimension to form a six-channel input, enabling the model to analyse pre- and post-disaster visual information jointly. A pretrained Vision Transformer is used as a fixed feature extractor. The backbone is modified to accept six-channel inputs and is subsequently frozen in all experiments, while only the classification head is optimised. This design isolates the contribution of the head architecture and enables a controlled comparison among conventional, attention-based, geometry-aware, quantum-inspired, and hybrid quantum-classical token mixing strategies under an identical visual representation.

Five classification heads are evaluated using the same data splits, input preprocessing, frozen backbone, optimisation objective, and model-selection protocol: a multilayer perceptron (MLP) head, a shallow Transformer head, a Grassmann–Plücker (GP) head, a Quantum-inspired Grassmann–Plücker (QGP) head, and a Hybrid Quantum Machine Learning Grassmann–Plücker (HQML-GP) head. The task is formulated as building-centred four-class damage classification, rather than full-scene building localisation or semantic segmentation. This experimental design therefore focuses specifically on how different head-level token mixing mechanisms use a shared frozen ViT representation for damage classification.

### 3.2. Datasets and event selection

The experiments use building-centred paired image patches derived from tornado-related events in the xBD dataset. Each sample contains a pre-event RGB image patch, a corresponding post-event RGB image patch, and a building damage label. The complete dataset is defined as (Eq. 1):

$$\mathcal{D} = \{(B_i, A_i, y_i)\}_{i=1}^{N} \tag{1}$$

Where $B_i$ and $A_i$denote the pre-event and post-event image patches for sample $i$, respectively, $y_i$ denotes the associated damage class, and $N$is the total number of samples. The classification label belongs to the set $y_i \in \mathcal{Y} = \{0,1,2,3\}$, with the class mapping 0 = No damage, 1 = Minor damage, 2 = Major damage, 3 = Destroyed. The study uses xBD case IDs 5 and 7, corresponding to the Joplin and Moore tornadoes, for model development and within-event evaluation. Samples from these two events are used to construct the training, validation, and seen-event test sets. Case ID 11, corresponding to the Tuscaloosa tornado, is excluded from all model-development stages and reserved exclusively for unseen-event testing.

This event-based design supports two complementary forms of evaluation. The seen-event test set assesses generalisation to held-out samples from disasters represented during training, whereas the unseen-event test set evaluates cross-event transfer to a tornado event not observed during model development. This distinction is important because satellite image characteristics, building typologies, damage patterns, and environmental context may differ across disaster events.

***Table 1**. Dataset split and class distribution.*

| Split | Events | No damage | Minor | Major | Destroyed | Total |
|---|---|---|---|---|---|---|
| Source train/validation pool | Joplin and Moore | 16,092 | 1,983 | 968 | 2,669 | 21,712 |
| Balanced training set | Joplin and Moore | 823 | 823 | 823 | 823 | 3,292 |
| Validation set | Joplin and Moore | 2,413 | 297 | 145 | 400 | 3,255 |
| Seen-disaster test set | Joplin and Moore | 3,549 | 412 | 186 | 507 | 4,654 |
| Unseen-disaster test set | Tuscaloosa | 1,407 | 282 | 75 | 123 | 1,887 |

Table 1 summarises the class distributions for the source pool, balanced training set, validation set, seen-event test set, and unseen-event test set. The validation and test sets retain their natural class distributions. Only the training set is balanced, allowing the models to be optimised without dominance by the majority no-damage class while preserving realistic class frequencies during model selection and final evaluation.

### 3.3. Data preparation and class balancing

Each pre-event image $B_i$ and corresponding post-event image $A_i$ is resized to $224 \times 224$ pixels using bilinear interpolation and converted to a three-channel tensor. The paired images are then concatenated along the channel dimension to form a six-channel input tensor (Eq. 2):

$$X_i = \mathrm{Concat}_{\mathrm{ch}}(B_i, A_i) \in \mathbb{R}^{6\times224\times224} \tag{2}$$

Where $\mathrm{Concat}_{\mathrm{ch}}$ denotes channel-wise concatenation. The first three channels correspond to the pre-event image and the remaining three channels correspond to the post-event image. The six-channel tensor is normalised using the ImageNet channel-wise mean and standard deviation, repeated for the pre-event and post-event channel groups (Eq. 3):

$$\begin{aligned} \boldsymbol{\mu} &= [0.485,\ 0.456,\ 0.406,\ 0.485,\ 0.456,\ 0.406] \\ \boldsymbol{\sigma} &= [0.229,\ 0.224,\ 0.225,\ 0.229,\ 0.224,\ 0.225] \end{aligned} \tag{3}$$

The normalized value at channel c and spatial location $(h, w)$ is calculated as (Eq. 4):

$$X'_{i,c,h,w} = \frac{X_{i,c,h,w} - \mu_c}{\sigma_c}, \qquad c \in \{1, \dots, 6\} \tag{4}$$

No data augmentation is applied in the final experimental protocol. This decision maintains an identical input-processing procedure across all classification heads and avoids introducing augmentation-related variability into the controlled head-level comparison.

The combined Joplin and Moore development pool is divided into training and validation subsets using class-stratified sampling. For each damage class, 15% of the available samples are assigned to the validation set, while the remaining 85% form the initial unbalanced training pool. The validation set retains the natural class distribution and is used exclusively for model selection and early stopping. Let (Eq. 5):

$$\mathcal{D}_{\text{train},c}^{\text{raw}} = \{(B_i, A_i, y_i) \in \mathcal{D}_{\text{train}}^{\text{raw}} \mid y_i = c\} \tag{5}$$

$\mathcal{D}_{\text{train},c}^{\text{raw}}$ denotes the subset of initial training samples belonging to class $c$, and let (Eq. 6):

$$n_c = \left|\mathcal{D}_{\text{train},c}^{\text{raw}}\right| \tag{6}$$

Represent the number of available samples in that class. The target number of samples retained from each class is defined by the size of the smallest class (Eq. 7):

$$n^* = \min_{c \in \mathcal{Y}} n_c \tag{7}$$

In the present dataset, major damage is the smallest class after validation splitting, giving $n^* = 823$. A balanced training set is therefore constructed by randomly selecting $823$ samples without replacement from each class (Eq. 8):

$$\mathcal{D}_{\text{train}}^{\text{bal}} = \bigcup_{c \in \mathcal{Y}} \mathcal{S}_c, \qquad \mathcal{S}_c \subseteq \mathcal{D}_{\text{train},c}^{\text{raw}}, \qquad |\mathcal{S}_c| = n^* \tag{8}$$

Only the training set is balanced. The validation, seen-event test, and unseen-event test sets retain their natural class distributions so that model selection and final performance evaluation reflect realistic post-disaster damage frequencies. The stratified splitting and random under sampling procedures are performed using a fixed random seed of 42 to ensure reproducibility.

### 3.4. Frozen six-channel ViT-B/16 encoder

The visual encoder is a ViT-B/16 model pretrained on ImageNet-21k and obtained from the Hugging Face checkpoint *google/vit-base-patch16-224-in21k [64]*. The original ViT-B/16 patch-projection layer accepts three-channel RGB images and uses a convolution with kernel size and stride equal to the patch size P=16. Because the present study jointly processes paired pre- and post-event RGB images, the original three-channel projection is replaced with a six-channel convolution while preserving the original output dimension, kernel size, stride, padding, and bias configuration. Let: $W_3 \in R^{DViT \times 3 \times P \times P}$ denote the pretrained three-channel patch-projection weights and let: $W_6 \in R^{DViT \times 6 \times P \times P}$ denote the corresponding six-channel weights. The new projection weights are initialized by duplicating the pretrained RGB weights across the pre-event and post-event channel groups and scaling both copies by one half (Eq. 9):

$$W_6 = \frac{1}{2}\text{Concat}_{\text{in}}(W_3, W_3) \tag{9}$$

where $\text{Concat}_{\text{in}}$ denotes concatenation along the input-channel dimension. Equivalently, the first three input channels of $W_6$ process the pre-event image, while the remaining three channels process the post-event image. This factor $1/2$ prevents the duplicated projection from approximately doubling the activation magnitude relative to the original pretrained layer. If the pretrained projection contains a bias vector $b_3$, it is transferred directly to the new projection: $b_6 = b_3$. Following this modification, the six-channel patch projection and all remaining ViT parameters are frozen. The encoder therefore serves as a fixed visual feature extractor, and only the downstream classification head is optimised during training.

For each normalised six-channel input $X_i'$, the frozen encoder produces a sequence containing one classification token and $N_p$ patch tokens (Eq. 10):

$$H_i = f_{\text{ViT}}^{\text{frozen}}(X'_i) = \left[h_i^{\text{cls}}; T_i\right] \in \mathbb{R}^{(N_p+1) \times D_{\text{ViT}}} \tag{10}$$

Where $h_i^{\text{cls}} \in \mathbb{R}^{D_{\text{ViT}}}$ denotes the classification-token representation and $T_i = \left[t_{i,1}, t_{i,2}, \dots, t_{i,N_p}\right]^{\top} \in \mathbb{R}^{N_p \times D_{\text{ViT}}}$ denotes the sequence of patch token representations supplied to the trainable classification head.

For an input resolution of $224 \times 224$ pixels and a patch size of $16 \times 16$ pixels, the number of patch tokens is $N_p = (\frac{224}{16})^2 = 196$. and the ViT hidden dimension is $D_{\text{ViT}} = 768$. The classification token is excluded because all compared heads operate exclusively on the common sequence of 196 patch tokens. This frozen-backbone design ensures that performance differences arise from the token mixing and classification heads rather than from changes to the visual encoder.

### 3.5. Classification head architectures

All five classification heads receive the same sequence of patch token representations produced by the frozen six-channel ViT-B/16 encoder. $Ti \in R^{Np \times DViT}$, where $N_p = 196$ and $D_{\text{ViT}} = 768$. The heads differ only in how they transform, mix, and aggregate these patch tokens before damage classification. The ViT backbone remains frozen in all experiments, and only the parameters of the selected classification head are optimized. Let $\mathcal{M}$ denote the set of evaluated head architectures: $\mathcal{M} = \{\text{MLP}, \text{Transformer}, \text{GP}, \text{QGP}, \text{HQML-GP}\}$. For a head architecture $m \in \mathcal{M}$, the corresponding trainable mapping $f_{\theta_m}^{(m)}$ transforms the common ViT token sequence into a vector of class logits (Eq. 11):

$$\boldsymbol{\ell}_i^{(m)} = f_{\theta_m}^{(m)}(T_i) \in \mathbb{R}^C, \qquad m \in \mathcal{M}, \quad C = 4 \tag{11}$$

Where $\theta_m$ denotes the trainable parameters of head $m$, and $\boldsymbol{\ell}_i^{(m)}$ contains the logits associated with the four damage classes. The predicted class is obtained as (Eq. 12):

$$\hat{y}_i^{(m)} = \underset{c \in \{0,\dots,C-1\}}{\arg\max} \ \ell_{i,c}^{(m)} \tag{12}$$

The MLP head serves as a simple mean pooling baseline and does not explicitly model interactions between patch tokens. The Transformer head provides an attention-based token mixing baseline. The Grassmann–Plücker head is the principal proposed geometry-aware architecture and represents relationships between token pairs through Plücker coordinate subspace descriptors. The QGP head extends the GP representation using fully classical amplitude-derived probability features, whereas the HQML-GP head augments the GP representation with expectation values generated by a differentiable quantum circuit.

Accordingly, QGP and HQML-GP should be interpreted as extensions of the standard GP framework rather than as independent alternatives to it. Table 2 summarises the role and principal operation of each classification head.

*Table 2. Classification heads compared in the study.*

| Head | Principal operation | Role in the comparison |
|---|---|---|
| MLP | Mean-pools the ViT patch tokens and applies a feed-forward classifier | Simple non-token-mixing baseline |
| Transformer | Projects the patch tokens and applies a shallow Transformer encoder before mean pooling | Attention-based token mixing baseline |
| GP | Constructs Grassmann–Plücker representations from token pairs at multiple predefined offsets | Proposed geometry-aware head |
| QGP | Enriches the GP Plücker coordinates with amplitude-derived probability features | Fully classical quantum-inspired GP extension |
| HQML-GP | Combines local GP features with expectation values obtained from a simulated differentiable quantum circuit | Hybrid quantum-classical GP extension |

### 3.5.1. MLP and Transformer baseline heads

The MLP head provides a simple non-token-mixing baseline. Given the frozen ViT patch token sequence $T_i$, the tokens are first aggregated by global mean pooling (Eq. 13) [65,66]:

$$\bar{\mathbf{t}}_i = \frac{1}{N_p} \sum_{t=1}^{N_p} \mathbf{t}_{i,t} \in \mathbb{R}^{D_{\text{ViT}}} \tag{13}$$

Where $\mathbf{t}_{i,t} \in \mathbb{R}^{D_{\text{ViT}}}$ denotes the representation of patch token $t$ for sample $i$. The pooled representation is subsequently processed by layer normalisation, a linear hidden layer, GELU activation, dropout, and a final linear classifier (Eq. 14):

$$\mathbf{h}_i^{\mathrm{MLP}} = \mathrm{Dropout}(\mathrm{GELU}(W_1\,\mathrm{LN}(\bar{\mathbf{t}}_i) + \mathbf{b}_1))$$
$$\boldsymbol{\ell}_i^{\mathrm{MLP}} = W_2\mathbf{h}_i^{\mathrm{MLP}} + \mathbf{b}_2 \in \mathbb{R}^C \tag{14}$$

Where $W_1 \in \mathbb{R}^{D_{\mathrm{MLP}} \times D_{\mathrm{ViT}}}$, $W_2 \in \mathbb{R}^{C \times D_{\mathrm{MLP}}}$, $D_{\mathrm{MLP}} = 256$, and $C = 4$. A dropout rate of 0.15 is used. Because token aggregation is performed before the feed-forward transformation, this head does not explicitly model interactions between individual patch tokens.

The Transformer head provides an attention-based token mixing baseline [52,53]. The frozen ViT patch tokens are first linearly projected from $D_{\mathrm{ViT}} = 768$ to the common model dimension $D_{\mathrm{model}} = 256$ (Eq. 15):

$$Z_i^{(0)} = T_i W_{\mathrm{in}} + \mathbf{1}_{N_p}\mathbf{b}_{\mathrm{in}}^{\mathrm{T}} \in \mathbb{R}^{N_p \times D_{\mathrm{model}}} \tag{15}$$

Where $W_{\mathrm{in}} \in \mathbb{R}^{D_{\mathrm{ViT}} \times D_{\mathrm{model}}}$, $\mathbf{b}_{\mathrm{in}} \in \mathbb{R}^{D_{\mathrm{model}}}$, and $\mathbf{1}_{N_p}$ is an $N_p$-dimensional vector of ones used to broadcast the bias across all patch tokens. The projected sequence is then processed by a single Transformer encoder layer (Eq. 16):

$$Z_i^{(1)} = \mathrm{TransformerEncoder}_{\phi}\left(Z_i^{(0)}\right) \in \mathbb{R}^{N_p \times D_{\mathrm{model}}} \tag{16}$$

Where $\phi$ denotes the trainable parameters of the Transformer encoder. The encoder uses four self-attention heads, a feed-forward dimension of $512$, GELU activation, a dropout rate of 0.15, and batch-first processing. The resulting token representations are mean-pooled and passed through layer normalisation, dropout, and a final linear classifier (Eq. 17):

$$\bar{\mathbf{z}}_i = \frac{1}{N_p}\sum_{t=1}^{N_p} \mathbf{z}_{i,t}^{(1)} \in \mathbb{R}^{D_{\mathrm{model}}}$$
$$\boldsymbol{\ell}_i^{\mathrm{Tr}} = W_{\mathrm{c}}\,\mathrm{Dropout}(\mathrm{LN}(\bar{\mathbf{z}}_i)) + \mathbf{b}_{\mathrm{c}} \in \mathbb{R}^C \tag{17}$$

Where $W_c \in \mathbb{R}^{C \times D_{\text{model}}}$ and $\mathbf{b}_c \in \mathbb{R}^C$. This head evaluates whether conventional self-attention-based token mixing improves classification relative to simple mean pooling under the same frozen-backbone protocol.

### 3.5.2. Grassmann–Plücker head

The Grassmann–Plücker (GP) head is designed to represent relational information between patch tokens through geometric subspace descriptors [25,67]. Given the frozen ViT token sequence $T_i \in \mathbb{R}^{N_p \times D_{\text{ViT}}}$, each token is first projected from $D_{\text{ViT}} = 768$ to the common model dimension $D_{\text{model}} = 256$ (Eq. 18):

$$H_i = T_i W_{\text{in}}^{\text{GP}} + \mathbf{1}_{N_p} \left(\mathbf{b}_{\text{in}}^{\text{GP}}\right)^{\mathsf{T}} \in \mathbb{R}^{N_p \times D_{\text{model}}} \tag{18}$$

Where $W_{\text{in}}^{\text{GP}} \in \mathbb{R}^{D_{\text{ViT}} \times D_{\text{model}}}$, $\mathbf{b}_{\text{in}}^{\text{GP}} \in \mathbb{R}^{D_{\text{model}}}$, and $\mathbf{1}_{N_p}$ broadcasts the bias across all token positions. Within the GP mixing layer, the projected tokens are reduced to $D_{\text{proj}} = 64$ dimensions (Eq. 19):

$$Z_i = H_i W_{\text{red}} + \mathbf{1}_{N_p} \mathbf{b}_{\text{red}}^{\mathsf{T}} \in \mathbb{R}^{N_p \times D_{\text{proj}}} \tag{19}$$

Where $W_{\text{red}} \in \mathbb{R}^{D_{\text{model}} \times D_{\text{proj}}}$. The reduced representation is divided into $H = 4$ heads (Eq. 20):

$$Z_i = \text{Concat}\left(Z_i^{(1)}, \dots, Z_i^{(H)}\right), \qquad Z_i^{(h)} \in \mathbb{R}^r \tag{20}$$

Where $r = \frac{D_{\text{proj}}}{H} = \frac{64}{4} = 16$. Token pairs are constructed using the predefined offset set: $\mathcal{K} = \{1,2,4,8,14,28\}$. For token position t, the set of valid forward offsets is defined as (Eq. 21):

$$\mathcal{K}_t = \left\{k \in \mathcal{K} \mid t + k \le N_p\right\} \tag{21}$$

For each head $h$, token position $t$, and valid offset $k \in \mathcal{K}_t$, the ordered pair $\left(\mathbf{z}_{i,t}^{(h)}, \mathbf{z}_{i,t+k}^{(h)}\right)$ is represented using its exterior product. When the two vectors are linearly independent, this representation corresponds to the two-dimensional subspace they span. The associated Plücker coordinate vector is defined component-wise as (Eq. 22):

$$\left[\mathbf{p}_{i,t,k}^{(h)}\right]_{ab} = z_{i,t,a}^{(h)} z_{i,t+k,b}^{(h)} - z_{i,t,b}^{(h)} z_{i,t+k,a}^{(h)}, \qquad 1 \le a < b \le r \tag{22}$$

The number of distinct Plücker coordinates produced by each head is $D_{\text{Pl}} = \binom{r}{2} = \frac{r(r-1)}{2} = 120$. To reduce sensitivity to the magnitude of the two input vectors and improve numerical stability, each Plücker vector is $L_2$-normalised (Eq. 23):

$$\tilde{\mathbf{p}}_{i,t,k}^{(h)} = \frac{\mathbf{p}_{i,t,k}^{(h)}}{\max\left(\left\|\mathbf{p}_{i,t,k}^{(h)}\right\|_2, \varepsilon\right)} \tag{23}$$

Where $\varepsilon > 0$ is a small numerical-stability constant. Each normalised Plücker vector is then projected to the per-head model dimension $d_h = D_{\text{model}}/H = 64$ (Eq. 24):

$$\mathbf{s}_{i,t,k}^{(h)} = W_{\text{Pl}}^{(h)} \tilde{\mathbf{p}}_{i,t,k}^{(h)} + \mathbf{b}_{\text{Pl}}^{(h)} \in \mathbb{R}^{d_h} \tag{24}$$

Where $W_{\text{Pl}}^{(h)} \in \mathbb{R}^{d_h \times D_{\text{Pl}}}$. The geometric representations obtained from all valid offsets are averaged for each token and head (Eq. 25):

$$\mathbf{g}_{i,t}^{(h)} = \frac{1}{|\mathcal{K}_t|} \sum_{k \in \mathcal{K}_t} \mathbf{s}_{i,t,k}^{(h)} \tag{25}$$

When $\mathcal{K}_t$ is empty, the corresponding sum is defined as the zero vector. The head-specific geometric representations are subsequently concatenated (Eq. 26):

$$\mathbf{g}_{i,t} = \text{Concat}\left(\mathbf{g}_{i,t}^{(1)}, \dots, \mathbf{g}_{i,t}^{(H)}\right) \in \mathbb{R}^{D_{\text{model}}} \tag{26}$$

The geometric representation $\mathbf{g}_{i,t}$ is fused with the original projected token representation $\mathbf{h}_{i,t}$ using a learnable element-wise gate (Eq. 27):

$$\begin{gathered} \boldsymbol{\alpha}_{i,t} = \sigma\left(W_g\left[\mathbf{h}_{i,t}; \mathbf{g}_{i,t}\right] + \mathbf{b}_g\right) \in (0,1)^{D_{\text{model}}} \\ \mathbf{m}_{i,t} = \boldsymbol{\alpha}_{i,t} \odot \mathbf{h}_{i,t} + \left(\mathbf{1} - \boldsymbol{\alpha}_{i,t}\right) \odot \mathbf{g}_{i,t} \end{gathered} \tag{27}$$

Where $[\cdot\,;\cdot]$ denotes vector concatenation, $\sigma(\cdot)$ is the element-wise sigmoid function, and $\odot$ denotes element-wise multiplication. The gate allows the model to adaptively balance the original token representation and its GP-derived geometric feature. The fused token is processed using a residual feed-forward transformation (Eq. 28, 29):

$$\mathbf{u}_{i,t} = \mathrm{LN}\big(\mathbf{m}_{i,t} + \mathrm{Dropout}\big(\mathrm{FFN}(\mathbf{m}_{i,t})\big)\big) \tag{28}$$

Where:

$$\mathrm{FFN}(\mathbf{x}) = W_{\mathrm{ff},2}\,\mathrm{GELU}\big(W_{\mathrm{ff},1}\mathbf{x} + \mathbf{b}_{\mathrm{ff},1}\big) + \mathbf{b}_{\mathrm{ff},2} \tag{29}$$

The resulting sequence is mean-pooled and passed to the final classifier (Eq. 30):

$$\begin{gathered}\bar{\mathbf{u}}_i = \frac{1}{N_p}\sum_{t=1}^{N_p}\mathbf{u}_{i,t} \in \mathbb{R}^{D_{\mathrm{model}}} \\ \boldsymbol{\ell}_i^{\mathrm{GP}} = W_{\mathrm{out},2}\,\mathrm{Dropout}\big(\mathrm{GELU}\big(W_{\mathrm{out},1}\bar{\mathbf{u}}_i + \mathbf{b}_{\mathrm{out},1}\big)\big) + \mathbf{b}_{\mathrm{out},2} \in \mathbb{R}^{C}\end{gathered} \tag{30}$$

The classifier uses a hidden dimension of 512, a GELU activation, a dropout rate of 0.15, and $C = 4$ output classes. This architecture enables the GP head to combine the original ViT token information with explicit geometric representations of token pair relationships while retaining the common frozen-backbone comparison protocol.

### 3.5.3. Quantum-inspired Grassmann–Plücker head

The Quantum-inspired Grassmann–Plücker (QGP) head extends the standard GP head by enriching each normalized Plücker coordinate vector with amplitude-derived squared features. The complete computation remains classical and does not involve a quantum circuit or quantum hardware [67,68]. The input projection, dimensionality reduction, head partitioning, token pair construction, and Plücker coordinate computation are identical to those defined for the GP head. For each sample $i$, token position $t$, offset $k \in \mathcal{K}_t$, and head $h$, let: $\tilde{\mathbf{p}}_{i,t,k}^{(h)} \in \mathbb{R}^{D_{\mathrm{Pl}}}$ denote the $L_2$-normalised Plücker vector, where $D_{\mathrm{Pl}} = 120$. This vector is first mapped to a lower-dimensional amplitude representation (Eq. 31):

$$\mathbf{a}_{i,t,k}^{(h)} = W_q^{(h)} \tilde{\mathbf{p}}_{i,t,k}^{(h)} + \mathbf{b}_q^{(h)} \in \mathbb{R}^{D_q} \tag{31}$$

Where $W_q^{(h)} \in \mathbb{R}^{D_q \times D_{\text{Pl}}}$, $\mathbf{b}_q^{(h)} \in \mathbb{R}^{D_q}$, and $D_q = 8$. The resulting vector is $L_2$-normalised (Eq. 32):

$$\boldsymbol{\psi}_{i,t,k}^{(h)} = \frac{\mathbf{a}_{i,t,k}^{(h)}}{\max\left(\left\|\mathbf{a}_{i,t,k}^{(h)}\right\|_2, \varepsilon\right)} \in \mathbb{R}^{D_q} \tag{32}$$

Where $\varepsilon > 0$ is a numerical-stability constant. A trainable linear transformation without bias is then applied to mix the amplitude components (Eq. 33):

$$\mathbf{v}_{i,t,k}^{(h)} = M_q^{(h)} \boldsymbol{\psi}_{i,t,k}^{(h)} \in \mathbb{R}^{D_q} \tag{33}$$

Where $M_q^{(h)} \in \mathbb{R}^{D_q \times D_q}$. The mixed vector is normalised again (Eq. 34):

$$\boldsymbol{\phi}_{i,t,k}^{(h)} = \frac{\mathbf{v}_{i,t,k}^{(h)}}{\max\left(\left\|\mathbf{v}_{i,t,k}^{(h)}\right\|_2, \varepsilon\right)} \in \mathbb{R}^{D_q} \tag{34}$$

The element-wise squared amplitudes are subsequently scaled by a trainable scalar parameter $\gamma_q$, initialised to 0.03 (Eq. 35):

$$\mathbf{q}_{i,t,k}^{(h)} = \gamma_q \left(\boldsymbol{\phi}_{i,t,k}^{(h)} \odot \boldsymbol{\phi}_{i,t,k}^{(h)}\right) \in \mathbb{R}^{D_q} \tag{35}$$

Before application of the scale parameter, the squared components are non-negative and sum to one whenever the normalisation denominator is determined by the vector norm rather than by $\varepsilon$. They are therefore interpreted as probability-like features rather than physical quantum measurement probabilities. The original normalised Plücker vector and the amplitude-derived feature vector are concatenated and projected to the per-head output dimension $d_h = 64$ (Eq. 36, 37):

$$\mathbf{s}_{i,t,k}^{\text{QGP},(h)} = W_{\text{pq}}^{(h)} \left[\tilde{\mathbf{p}}_{i,t,k}^{(h)}; \mathbf{q}_{i,t,k}^{(h)}\right] + \mathbf{b}_{\text{pq}}^{(h)} \in \mathbb{R}^{d_h} \tag{36}$$

Where:

$$W_{\text{pq}}^{(h)} \in \mathbb{R}^{d_h \times (D_{\text{Pl}} + D_q)}, \; and, \qquad \mathbf{b}_{\text{pq}}^{(h)} \in \mathbb{R}^{d_h} \tag{37}$$

Because $D_{\mathrm{Pl}} = 120$ and $D_q = 8$, the concatenated representation has dimension 128. The QGP features are averaged over the valid offsets for each token and head (Eq. 38):

$$\mathbf{g}_{i,t}^{\mathrm{QGP},(h)} = \frac{1}{|\mathcal{K}_t|} \sum_{k \in \mathcal{K}_t} \mathbf{s}_{i,t,k}^{\mathrm{QGP},(h)} \tag{38}$$

For token positions with no valid forward offset, the corresponding feature is defined as the zero vector. The head-specific representations are then concatenated (Eq. 39):

$$\mathbf{g}_{i,t}^{\mathrm{QGP}} = \mathrm{Concat}\left(\mathbf{g}_{i,t}^{\mathrm{QGP},(1)}, \dots, \mathbf{g}_{i,t}^{\mathrm{QGP},(H)}\right) \in \mathbb{R}^{D_{\mathrm{model}}} \tag{39}$$

The resulting QGP representation replaces the standard GP feature $\mathbf{g}_{i,t}$ in the gating, residual feed-forward, mean pooling, and classification operations. The QGP head therefore preserves the geometric token mixing structure of the standard GP architecture while introducing a compact nonlinear enrichment of each Plücker representation.

### 3.5.4. Hybrid quantum-classical Grassmann–Plücker head

The Hybrid Quantum Machine Learning Grassmann–Plücker (HQML-GP) head combines the local geometric token representations of the standard GP head with a global feature vector produced by a small differentiable quantum circuit [67,69]. The input projection, dimensionality reduction, head partitioning, token pair construction, Plücker coordinate computation, and local GP aggregation. The quantum component therefore supplements the GP representation rather than replacing it. For each offset $k \in \mathcal{K}$, define the set of token positions for which a valid forward pair exists as: $\mathcal{I}_k = \{t \in \{1, \dots, N_p\} \mid t + k \leq N_p\}$. For each sample $i$, head $h$, and offset $k$, the normalised Plücker coordinate vectors are averaged across all valid token positions (Eq. 40):

$$\bar{\mathbf{p}}_{i,k}^{(h)} = \frac{1}{|\mathcal{I}_k|} \sum_{t \in \mathcal{I}_k} \tilde{\mathbf{p}}_{i,t,k}^{(h)} \in \mathbb{R}^{D_{\mathrm{Pl}}} \tag{40}$$

The resulting vectors are concatenated across all offsets and heads to form a global Plücker descriptor (Eq. 41):

$$\mathbf{p}_i^{\text{global}} = \text{Concat}_{k\in\mathcal{K}}\left(\text{Concat}_{h=1}^{H}\left(\overline{\mathbf{p}}_{i,k}^{(h)}\right)\right) \in \mathbb{R}^{|\mathcal{K}|HD_{\text{Pl}}} \tag{41}$$

With $|\mathcal{K}| = 6$, $H = 4$, and $D_{\text{Pl}} = 120$, the global descriptor dimension is: $|\mathcal{K}|HD_{\text{Pl}} = 6 \times 4 \times 120 = 2880$. The global descriptor is mapped to $Q = 4$ bounded rotation angles (Eq. 42, 43):

$$\boldsymbol{\theta}_i = \pi \tanh\left(W_\theta \mathbf{p}_i^{\text{global}} + \mathbf{b}_\theta\right) \in [-\pi, \pi]^Q \tag{42}$$

Where:

$$W_\theta \in \mathbb{R}^{Q\times 2880}, \qquad \mathbf{b}_\theta \in \mathbb{R}^Q \tag{43}$$

The quantum circuit is implemented using PennyLane with the *default.qubit* simulator and differentiable Torch backpropagation. The circuit contains $Q = 4$ qubits initialised in the computational basis state $|\psi_0\rangle = |0\rangle^{\otimes Q}$. The input angles are encoded using single-qubit $R_Y$ rotations. A single trainable variational layer subsequently applies one $R_Y$ rotation and one $R_Z$ rotation to each qubit, followed by a ring of controlled-NOT gates. The resulting quantum state can be written as (Eq. 44):

$$|\psi_i\rangle = U_{\text{ent}}\left[\bigotimes_{j=1}^{Q} R_Z(\omega_{j,2})R_Y(\omega_{j,1})\right]\left[\bigotimes_{j=1}^{Q} R_Y(\theta_{i,j})\right]|0\rangle^{\otimes Q} \tag{44}$$

Where $\omega_{j,1}$ and $\omega_{j,2}$ are trainable circuit parameters and $U_{\text{ent}}$ denotes the ring of CNOT gates connecting adjacent qubits, including the final connection from qubit $Q$ to qubit $1$. The circuit output is obtained by measuring the expectation value of the Pauli-$Z$ operator on each qubit (Eq. 45):

$$q_{i,j} = \langle\psi_i|Z_j|\psi_i\rangle, \qquad j = 1, \dots, Q \tag{45}$$

The quantum output vector is therefore (Eq. 46):

$$\mathbf{q}_i = \left[q_{i,1}, q_{i,2}, \dots, q_{i,Q}\right]^{\top} \in [-1,1]^Q \tag{46}$$

The four expectation values are projected to the common model dimension: $\mathbf{r}_i^{\mathrm{Q}} = W_{\mathrm{Q}}\mathbf{q}_i + \mathbf{b}_{\mathrm{Q}} \in \mathbb{R}^{D_{\mathrm{model}}}$ where $W_{\mathrm{Q}} \in \mathbb{R}^{D_{\mathrm{model}} \times Q}, \mathbf{b}_{\mathrm{Q}} \in \mathbb{R}^{D_{\mathrm{model}}}$. The projected quantum feature is broadcast across all token positions and added to the standard GP representation using a trainable scalar $\lambda_{\mathrm{Q}}$ (Eq. 47):

$$\mathbf{g}_{i,t}^{\mathrm{HQML}} = \mathbf{g}_{i,t}^{\mathrm{GP}} + \lambda_{\mathrm{Q}}\mathbf{r}_i^{\mathrm{Q}}, \qquad t = 1, \dots, N_p \tag{47}$$

The scale parameter is initialised as: $\lambda_{\mathrm{Q}}^{(0)} = 0.015$. The resulting hybrid representation replaces $\mathbf{g}_{i,t}$ in the gating, residual feed-forward, mean pooling, and classification operations. Because the same global quantum feature is supplied to every token, the HQML-GP head combines local token pair geometry with sample-level global quantum circuit information. The quantum circuit is evaluated using a classical simulator. The HQML-GP head is therefore used to investigate whether circuit-derived nonlinear global features improve the GP representation, rather than to claim quantum computational advantage.

### 3.6. Training, model selection, and implementation

All classification heads were trained using the same optimization protocol to ensure a controlled comparison. For a mini-batch of $B$ samples, the output logits of the selected head are converted to class probabilities using the SoftMax function (Eq. 48):

$$p_{i,c} = \frac{\exp(\ell_{i,c})}{\sum_{j=0}^{C-1} \exp(\ell_{i,j})}, \qquad c \in \{0, \dots, C-1\} \tag{48}$$

Where $\ell_{i,c}$ denotes the logit assigned to class $c$ for sample $i$, $p_{i,c}$ is the corresponding predicted probability, and $C = 4$. The trainable parameters of each classification head were optimised using the categorical cross-entropy loss (Eq. 49):

$$\mathcal{L}_{\mathrm{CE}} = -\frac{1}{B}\sum_{i=1}^{B} \log p_{i,y_i} \tag{49}$$

The AdamW optimiser was used for all heads with a learning rate of $5 \times 10^{-5}$ and a weight-decay coefficient of $1 \times 10^{-4}$. The batch size was fixed at 16, and training was conducted for a maximum of 20 epochs. Gradient clipping was applied using a maximum global gradient norm of 1.0 to reduce the risk of unstable parameter updates. Early stopping was based on validation macro-F1 rather than validation accuracy because the validation set retained its naturally imbalanced class distribution. Let $F_{\text{macro}}^{\text{val}}(e)$ denote the validation macro-F1 obtained at epoch $e$. The selected checkpoint was defined as (Eq. 50):

$$e^* = \arg\max_{1 \le e \le E} F_{\text{macro}}^{\text{val}}(e) \tag{50}$$

Where $E \le 20$ is the number of epochs completed before early stopping. A patience of five epochs was used for the heads. The checkpoint corresponding to $e^*$ was subsequently evaluated on the seen-event and unseen-event test sets. The full experimental configuration is presented in Table 3, and the corresponding model-architecture settings are provided in Table 4. Automatic mixed precision was used for the classical heads but disabled for HQML-GP because its differentiable quantum circuit was evaluated through the PennyLane *default.qubit* simulator.

*Table 3. Main experimental configuration.*

| **Component** | **Setting** |
| --- | --- |
| Input size | 224×224 |
| Input channels | 6 |
| Damage classes | 4 |
| Backbone | ViT-B/16 pretrained on ImageNet-21k |
| Hugging Face model | *google/vit-base-patch16-224-in21k* |
| Patch size | 16×16 |
| Patch tokens | 196 |
| ViT hidden dimension | 768 |
| Frozen parameters | All ViT backbone parameters |
| Trainable parameters | Classification head only |
| Batch size | 16 |
| Optimizer | *AdamW* |
| Learning rate | $5 \times 10^{-5}$ |
| Weight decay | $1 \times 10^{-4}$ |
| Maximum epochs | 20 |
| Early stopping metric | Validation macro-F1 |
| Patience | 5 epochs |
| Dropout | 0.15 |
| Gradient clipping | Maximum norm 1.0 |
| Random seed | 42 |
| Data augmentation | None |
| Classical AMP | Enabled |
| Hybrid AMP | Disabled |
| Hardware | NVIDIA GeForce RTX 3090 |

*Table 4. Head architecture configuration.*

| **Hyperparameter** | **Value** |
|---|---|
| $D_{\mathrm{ViT}}$ | 768 |
| $D_{\mathrm{model}}$ | 256 |
| $D_{\mathrm{FF}}$ | 512 |
| $D_{\mathrm{proj}}$ | 64 |
| Number of heads | 4 |
| Number of GP/Transformer layers | 1 |
| GP offsets | 1, 2, 4, 8, 14, 28 |
| Reduced dimension per GP head | 16 |
| Plucker coordinates per head | 120 |
| QGP amplitude dimension per head | 8 |
| QGP scale initialization | 0.03 |
| HQML-GP qubits | 4 |
| HQML-GP quantum layers | 1 |
| HQML-GP global Plucker descriptor dimension | 2880 |
| HQML-GP quantum output dimension | 4 |

| HQML-GP quantum scale initialization | 0.015 |
|---|---|

### 3.7. Evaluation metrics

The checkpoint selected according to validation macro-F1 was evaluated independently on the seen-event and unseen-event test sets. Predictive performance was assessed using accuracy, balanced accuracy, macro precision, macro recall, class-specific F1, macro-F1, weighted-F1, and the multiclass Matthews correlation coefficient (MCC) [50,70,71]. Trainable parameter count, average training time per epoch, and inference time per image were also reported to compare computational efficiency [72,73]. For each damage class $c$, precision, recall, and F1 were calculated as (Eq. 51):

$$\text{Precision}_c = \frac{TP_c}{TP_c + FP_c}, \qquad \text{Recall}_c = \frac{TP_c}{TP_c + FN_c},$$
$$F1_c = \frac{2TP_c}{2TP_c + FP_c + FN_c} \tag{51}$$

where $TP_c$, $FP_c$, and $FN_c$ denote the true-positive, false-positive, and false-negative counts for class $c$, respectively. Overall accuracy was calculated as (Eq. 52):

$$\text{Accuracy} = \frac{\sum_{c=0}^{C-1} TP_c}{N} \tag{52}$$

Where $C = 4$ is the number of damage classes and $N$ is the total number of evaluated samples.

Balanced accuracy was calculated as the unweighted mean of the class-specific recall values (Eq. 53):

$$\text{Balanced Accuracy} = \frac{1}{C}\sum_{c=0}^{C-1} \text{Recall}_c \tag{53}$$

Macro precision and macro recall were obtained by averaging the corresponding class-specific values across the four damage classes. In single-label multiclass classification, macro recall is mathematically equivalent to balanced accuracy. Macro-F1 gives equal importance to all classes, whereas weighted-F1 accounts for differences in class support (Eq. 54):

$$\text{Macro-F1} = \frac{1}{C}\sum_{c=0}^{C-1} F\,1_c, \qquad \text{Weighted-F1} = \sum_{c=0}^{C-1} \frac{n_c}{N} F1_c \tag{54}$$

Where $n_c$ is the number of ground-truth samples belonging to class $c$. The multiclass MCC was calculated from the complete confusion matrix as (Eq. 55):

$$\text{MCC} = \frac{s\tau - \sum_{c=0}^{C-1} p_c\, t_c}{\sqrt{(s^2 - \sum_{c=0}^{C-1} p_c^2)(s^2 - \sum_{c=0}^{C-1} t_c^2)}} \tag{54}$$

Where $\tau$is the total number of correctly classified samples, $s = N$is the total number of samples, $p_c$ is the number of samples predicted as class $c$, and $t_c$is the number of samples whose ground-truth label is class $c$. MCC ranges from $-1$ to $1$, with larger values indicating stronger agreement between predicted and observed labels.

The number of trainable parameters includes only parameters updated during optimisation and excludes the frozen ViT backbone. Average training time per epoch and inference time in milliseconds per image were measured using the computational configuration described in Section 3.6.

# 4. Results

## 4.1. Overall performance, cross-event generalisation

Table 5 summarises the validation, test set, and computational results for the MLP, Transformer, GP, QGP, and HQML-GP classification heads. All models were evaluated using the same frozen six-channel ViT-B/16 backbone, data partitions, optimisation procedure, checkpoint selection criterion, and evaluation protocol. The comparison therefore examines how the classification head design influences predictive performance and computational requirements under a common visual representation.

The best validation macro-F1 corresponds to the highest validation macro-F1 obtained during training and was used for checkpoint selection and early stopping. The HQML-GP head achieved the highest validation macro-F1 point estimate of 65.63%, followed by the QGP head at 65.04%, the GP head at 64.59%, the MLP head at 64.06%, and the Transformer head at 63.56%. The difference between the two leading models was therefore 0.59 percentage points. Although the HQML-GP head achieved the highest validation value, this advantage did not translate into the highest performance on either test set.

On the seen disaster test set, the QGP head achieved the highest accuracy (83.46%), macro precision (62.22%), macro-F1 (64.50%), weighted F1 (84.82%), and MCC (62.52%). The HQML-GP head produced closely comparable results, with an accuracy of 82.49%, macro-F1 of 64.40%, weighted F1 of 84.14%, and MCC of 61.36%. The macro-F1 difference between these two heads was only 0.10 percentage points, indicating that their overall class-balanced test performance was very similar under the reported experimental configuration.

The MLP head achieved the highest seen-disaster balanced accuracy at 70.29%, narrowly exceeding the GP head at 70.21% and the HQML-GP head at 69.92%. The QGP head did not achieve the highest balanced accuracy but obtained the highest macro precision and macro-F1. The standard GP head also remained competitive with the conventional baselines, slightly exceeding the MLP head in accuracy, weighted F1, and MCC, while achieving a balanced accuracy that differed by only 0.08 percentage points. These results show that standard GP token mixing provided competitive head-level adaptation, although it did not consistently outperform all conventional baselines.

Performance decreased for all models when evaluated on the unseen Tuscaloosa tornado, demonstrating the difficulty of transfer across disaster events. However, the QGP head achieved the highest unseen-event accuracy (66.45%), macro-F1 (52.70%), weighted F1 (70.00%), and MCC (40.27%). However, its macro-F1 exceeded that of the Transformer head by only 0.20 percentage points, showing that the difference between the two leading models was modest. The HQML-GP head achieved the second-highest accuracy (64.28%) and MCC (39.67%), whereas the standard GP head produced results broadly comparable to the MLP and Transformer baselines.

The Transformer head achieved the highest unseen-event balanced accuracy (62.17%) and macro precision (51.82%), indicating that no single classification head dominated every evaluation criterion. The QGP head provided the strongest point estimates for several overall metrics, whereas the Transformer retained an advantage in average class recall and macro precision on the held-out event. Consequently, the results support a differentiated interpretation: quantum-inspired enrichment was associated with the strongest overall point estimates on the test sets, while conventional attention remained competitive for class-balanced recognition.

***Table 5**: Validation, predictive, and computational performance of the conventional and proposed classification heads on the seen-disaster test set comprising the Joplin and Moore tornadoes and the unseen-disaster test set comprising the Tuscaloosa tornado. The highest value for each predictive metric is shown in bold.*

| Model name | Train | Test - Seen disasters (Joplin and Moore tornadoes) | | | | | | | | | |
|---|---|---|---|---|---|---|---|---|---|---|---|
| | Best Val. Macro F1 | Accuracy | Balanced Accuracy | Macro Precision | Macro Recall | Macro F1 | Weighted F1 | MCC (%) | Params trainable | Time per epoch | ms per image |
| MLP head | 64.06 | 80.55 | **70.29** | 60.30 | **70.29** | 63.26 | 82.83 | 59.50 | 199,428 | 25.92s | 39.39 |
| Transformer attention head | 63.56 | 79.35 | 69.86 | 61.07 | 69.86 | 62.83 | 82.05 | 57.80 | 725,508 | 17.90s | 6.11 |
| GP head | 64.59 | 81.13 | 70.21 | 60.71 | 70.21 | 63.51 | 83.27 | 59.98 | 773,700 | 35.53s | 38.90 |
| Quantum-inspired GP head | 65.04 | **83.46** | 68.76 | **62.22** | 68.76 | **64.50** | **84.82** | **62.52** | 779,880 | 35.23s | 36.35 |
| Hybrid QML-GP head | **65.63** | 82.49 | 69.92 | 61.73 | 69.92 | 64.40 | 84.14 | 61.36 | 786,513 | 91.51s | 47.04 |

| Model name | Train | Test - Unseen disaster (Tuscaloosa tornado) | | | | | | | | | |
|---|---|---|---|---|---|---|---|---|---|---|---|
| | Best Val. Macro F1 | Accuracy | Balanced Accuracy | Macro Precision | Macro Recall | Macro F1 | Weighted F1 | MCC (%) | Params trainable | Time per epoch | ms per image |
| MLP head | - | 59.25 | 61.96 | 48.92 | 61.96 | 50.13 | 63.86 | 37.25 | 199,428 | 25.92s | 39.40 |
| Transformer attention head | - | 62.43 | **62.17** | **51.82** | **62.17** | 52.50 | 66.80 | 38.96 | 725,508 | 17.90s | 6.15 |
| GP head | - | 61.95 | 60.78 | 49.14 | 60.78 | 50.45 | 66.61 | 37.75 | 773,700 | 35.53s | 39.40 |
| Quantum-inspired GP head | - | **66.45** | 60.98 | 49.99 | 60.98 | **52.70** | **70.00** | **40.27** | 779,880 | 35.23s | 36.38 |
| Hybrid QML-GP head | - | 64.28 | 60.77 | 49.51 | 60.77 | 51.21 | 68.60 | 39.67 | 786,513 | 91.51s | 46.42 |

The MLP head contained the fewest trainable parameters, with 199,428 parameters, whereas the Transformer head contained 725,508 parameters and the QGP head contained 779,880 parameters. The Transformer head had the shortest reported training time per epoch at 17.90 s, followed by the MLP head at 25.92 s, the QGP head at 35.23 s, and the standard GP head at 35.53 s. The HQML-GP head required 91.51 s per epoch, reflecting the additional cost associated with the simulated differentiable quantum circuit. Under the timing procedure used in this study, the Transformer and QGP heads also recorded the lowest inference times, at approximately 6.1 and 6.4 ms per image, respectively. These timing values are implementation- and hardware-dependent and should therefore be interpreted together with the computational configuration reported in Section 3.6.

Overall, the results identify QGP as the strongest overall configuration achieving the highest point estimates for accuracy, macro-F1, weighted F1, and MCC on both the seen- and unseen-disaster test sets, while maintaining the second-lowest reported inference time. The HQML-GP head achieved the highest validation macro-F1 but did not surpass the QGP head on either test set or required substantially greater training time. The standard GP head remained competitive with the MLP and Transformer baselines but did not provide a consistent advantage across all metrics. Accordingly, under the fixed data split, random seed, and implementation used in this study, the results showed that the proposed quantum-inspired token mixer can use the same frozen visual representation more effectively than the matched Transformer attention head on several principal test metrics.

### 4.2. Class-specific performance and error analysis

Figure 2 presents representative pre- and post-disaster image pairs from the seen- and unseen-disaster test sets for the four damage classes. For each damage class and evaluation setting, the figure includes one sample correctly classified by all five heads and one sample misclassified by all five heads. The examples provide a qualitative illustration of damage patterns that produce either consistent agreement or shared difficulty across the evaluated classification heads.

The unanimously correct examples generally contain comparatively clear visual evidence. Destroyed buildings exhibit extensive structural loss, roof collapse, or widespread debris, whereas no-damage buildings retain visible structural continuity between the pre- and post-disaster images. The minor- and major-damage examples correctly classified by all five heads also contain relatively distinct changes in roof condition, debris distribution, or structural integrity. Agreement across all architectures in these cases indicates that the common frozen ViT representation contains sufficiently clear evidence for the downstream heads to reach the same correct decision.

The examples misclassified by all models tend to exhibit more visually ambiguous characteristics. In some cases, dust, debris, shadow, or changes in the surrounding area alter the post-disaster appearance even though damage to the target building is limited. Conversely, damage to a target structure may be visually subtle or partially obscured. These contextual and transitional patterns can lead the models to overestimate or underestimate the assigned damage severity. All five heads misclassify the selected examples, although they do not necessarily produce the same incorrect category in every case. The examples therefore illustrate shared failure conditions rather than proving that the underlying samples are intrinsically unclassifiable.

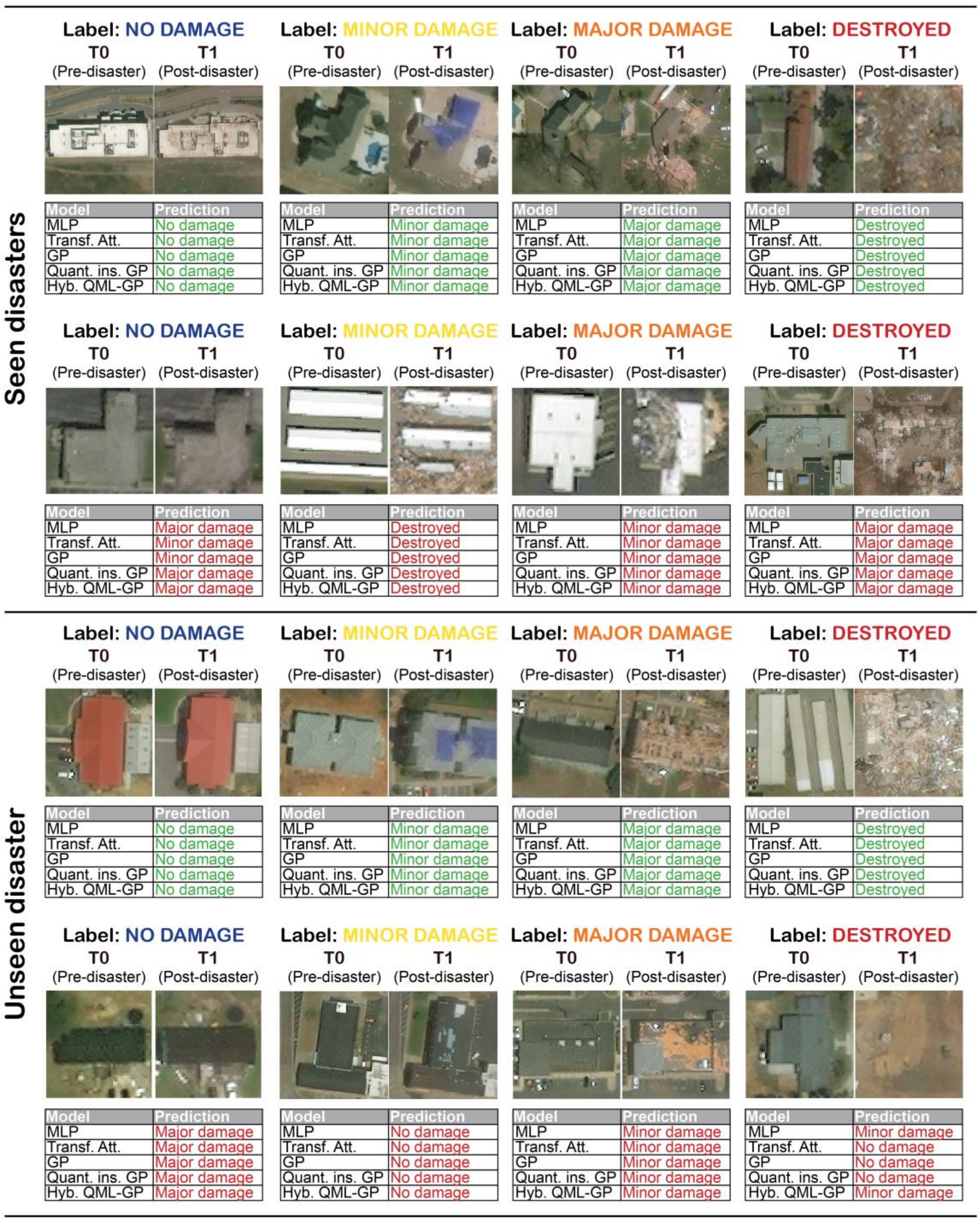

***Figure 2*. Representative pre- and post-disaster image pairs from the seen-event and unseen-event test sets across the four damage classes**. *For each damage class and evaluation setting, one sample correctly classified by all five heads and one sample misclassified by all five heads are shown. Green entries indicate correct predictions, whereas red entries indicate incorrect predictions. The examples illustrate cases of consistent model agreement and shared classification difficulty rather than the frequency of such outcomes in the test sets.*

The examples in Figure 2 are illustrative and should be interpreted together with the quantitative class-level results rather than as estimates of error frequency. Nevertheless, they indicate that remaining errors often involve intermediate or context-dependent damage patterns, where the visual condition of the target building may be difficult to separate from debris, damage, or surface changes in the surrounding area. Table 6 therefore examines whether these qualitative patterns are reflected in the class-specific F1 scores.

Table 6 reports the per-class F1 scores for the five classification heads on the seen- and unseen-disaster test sets. Across both evaluation settings, the no-damage and destroyed classes obtain substantially higher F1 scores than the minor- and major-damage classes. This pattern is common to all five architectures and indicates that the principal classification difficulty lies in distinguishing neighbouring intermediate damage states rather than separating visually intact buildings from severely destroyed structures. On the seen-disaster test set, the QGP head achieved the highest F1 scores for no damage at 92.4% and minor damage at 49.1%. The HQML-GP head achieved the highest destroyed-class F1 score at 77.9%, while the Transformer head achieved the highest major-damage F1 score at 42.1%. The standard GP head remained competitive but did not obtain the highest F1 score for any individual class. These results show that no single classification head dominated all four damage categories

***Table 6**: Per-class F1 scores (%) of the five classification heads on the seen-disaster and unseen-disaster test sets. Bold values identify the highest F1 score for each damage class and evaluation setting.*

| **Model name** | **Seen disasters** | | | | **Unseen disaster** | | | |
|---|---|---|---|---|---|---|---|---|
| | **No damage** | **Minor** | **Major** | **Destroyed** | **No damage** | **Minor** | **Major** | **Destroyed** |
| MLP head | 90.2 | 45.8 | 39.6 | 77.5 | 70.8 | 40.4 | 30.4 | 59.0 |
| Transformer attention head | 89.4 | 43.3 | **42.1** | 76.5 | 74.1 | 41.8 | **32.8** | **61.3** |
| GP head | 90.8 | 44.4 | 41.2 | 77.6 | 74.5 | 40.7 | 27.1 | 59.5 |
| Quantum-inspired GP head | **92.4** | **49.1** | 39.1 | 77.3 | **78.4** | 43.0 | 28.6 | 60.8 |
| Hybrid QML-GP head | 91.6 | 47.2 | 41.0 | **77.9** | 76.8 | **43.2** | 26.6 | 58.2 |

Class-specific performance declined when the models were transferred to the unseen Tuscaloosa event. The QGP head retained the highest no-damage F1 score at 78.4%, while the HQML-GP head achieved the highest minor-damage F1 score at 43.2%, narrowly exceeding the QGP head at 43.0%. The Transformer head achieved the highest unseen-event F1 scores for major damage at 32.8% and destroyed buildings at 61.3%. These results again show that the relative advantage of each head depends on the damage category considered.

Major damage remained the weakest class across the evaluated architectures. On the unseen event, major-damage F1 ranged from 26.6% for HQML-GP to 32.8% for the Transformer head. This class also showed a substantial reduction relative to seen-event performance for every architecture. The result is consistent with the intermediate visual position of major damage, which may overlap with both partial damage and complete destruction, and with its limited representation in the evaluation data. Consequently, improvements in aggregate accuracy or macro-F1 do not necessarily imply reliable recognition of operationally important intermediate damage states.

Figure 3 provides further detail by presenting row-normalised confusion matrices for the five heads on both test sets. As each row represents one true class and sums to 100%, the diagonal entries correspond to class-specific recall. Across the architectures, most errors occur between neighbouring severity categories. No-damage buildings are primarily confused with minor damage, whereas major-damage buildings are frequently confused with destroyed structures. Direct confusion between no damage and destruction is comparatively uncommon.

On the seen-disaster test set, the QGP head achieved the highest no-damage recall at 89.2%. Recognition of destroyed buildings was also comparatively strong across all architectures, with recall exceeding 75% for every evaluated head. These results are consistent with the relatively high no-damage and destroyed-class F1 scores reported in Table 6, although recall and F1 should not be interpreted as equivalent measures.

The minor- and major-damage categories remained more difficult. The standard GP head achieved the highest seen-event major-damage recall at 61.8%, narrowly exceeding the Transformer head at 61.3%. However, the Transformer achieved the highest major-damage F1 score in Table 6. This difference indicates that the Transformer produced a more favourable balance between precision and recall, despite having a slightly lower recall than the standard GP head.

**MLP head**

Seen disasters

| True \ Predicted | No | Minor | Major | Destroyed |
|---|---|---|---|---|
| No | 84.5% (2998) | 11.2% (396) | 2.5% (91) | 1.8% (64) |
| Minor | 21.1% (87) | 61.2% (252) | 15.0% (62) | 2.7% (11) |
| Major | 4.3% (8) | 18.3% (34) | 58.6% (109) | 18.8% (35) |
| Destroyed | 1.4% (8) | 1.6% (7) | 20.1% (102) | 76.9% (390) |

Unseen disaster

| True \ Predicted | No | Minor | Major | Destroyed |
|---|---|---|---|---|
| No | 56.7% (798) | 33.5% (471) | 6.5% (92) | 3.3% (46) |
| Minor | 14.2% (40) | 69.5% (196) | 11.0% (31) | 5.3% (15) |
| Major | 6.7% (5) | 17.3% (13) | 53.3% (40) | 22.7% (17) |
| Destroyed | 4.1% (5) | 7.3% (9) | 20.3% (25) | 68.3% (84) |

**Transformer attention head**

Seen disasters

| True \ Predicted | No | Minor | Major | Destroyed |
|---|---|---|---|---|
| No | 83.3% (2956) | 13.7% (486) | 1.9% (67) | 1.1% (40) |
| Minor | 20.6% (85) | 63.8% (263) | 13.7% (56) | 1.9% (8) |
| Major | 2.7% (5) | 22.0% (41) | 61.3% (114) | 14.0% (26) |
| Destroyed | 3.0% (15) | 2.5% (13) | 23.5% (119) | 71.0% (360) |

Unseen disaster

| True \ Predicted | No | Minor | Major | Destroyed |
|---|---|---|---|---|
| No | 61.4% (864) | 31.1% (437) | 5.6% (79) | 1.9% (27) |
| Minor | 17.0% (48) | 69.5% (196) | 10.3% (29) | 3.2% (9) |
| Major | 6.7% (5) | 20.0% (15) | 56.0% (42) | 17.3% (13) |
| Destroyed | 6.5% (8) | 6.5% (8) | 25.2% (31) | 61.8% (76) |

**GP head**

Seen disasters

| True \ Predicted | No | Minor | Major | Destroyed |
|---|---|---|---|---|
| No | 85.6% (3039) | 10.4% (369) | 2.5% (88) | 1.5% (53) |
| Minor | 22.3% (92) | 57.0% (235) | 17.0% (70) | 3.7% (15) |
| Major | 2.7% (5) | 16.7% (31) | 61.8% (115) | 18.8% (35) |
| Destroyed | 1.8% (9) | 2.4% (12) | 19.5% (99) | 76.3% (387) |

Unseen disaster

| True \ Predicted | No | Minor | Major | Destroyed |
|---|---|---|---|---|
| No | 62.0% (872) | 27.6% (388) | 7.7% (109) | 2.7% (38) |
| Minor | 17.4% (49) | 62.8% (177) | 15.2% (43) | 4.6% (13) |
| Major | 6.7% (5) | 20.0% (15) | 53.3% (40) | 20.0% (15) |
| Destroyed | 5.7% (7) | 6.5% (8) | 22.8% (28) | 65.0% (80) |

**Quantum-inspired GP head**

Seen disasters

| True \ Predicted | No | Minor | Major | Destroyed |
|---|---|---|---|---|
| No | 89.2% (3164) | 7.7% (275) | 1.7% (59) | 1.4% (51) |
| Minor | 26.5% (109) | 58.0% (239) | 12.6% (52) | 2.9% (12) |
| Major | 5.4% (10) | 21.4% (40) | 52.2% (97) | 21.0% (39) |
| Destroyed | 2.8% (14) | 1.4% (7) | 20.1% (102) | 75.7% (384) |

Unseen disaster

| True \ Predicted | No | Minor | Major | Destroyed |
|---|---|---|---|---|
| No | 68.3% (961) | 23.2% (326) | 5.5% (77) | 3.0% (43) |
| Minor | 24.1% (68) | 61.7% (174) | 9.2% (26) | 5.0% (14) |
| Major | 9.3% (7) | 24.0% (18) | 44.0% (33) | 22.7% (17) |
| Destroyed | 6.5% (8) | 7.3% (9) | 16.3% (20) | 69.9% (86) |

**Hybrid QML-GP head**

Seen disasters

| True \ Predicted | No | Minor | Major | Destroyed |
|---|---|---|---|---|
| No | 87.5% (3107) | 8.7% (309) | 2.2% (77) | 1.6% (56) |
| Minor | 26.0% (107) | 56.8% (234) | 14.8% (61) | 2.4% (10) |
| Major | 4.3% (8) | 17.2% (32) | 58.6% (109) | 19.9% (37) |
| Destroyed | 2.8% (14) | 1.0% (5) | 19.5% (99) | 76.7% (389) |

Unseen disaster

| True \ Predicted | No | Minor | Major | Destroyed |
|---|---|---|---|---|
| No | 65.2% (917) | 24.9% (350) | 7.3% (103) | 2.6% (37) |
| Minor | 18.1% (51) | 63.8% (180) | 13.5% (38) | 4.6% (13) |
| Major | 6.7% (5) | 20.0% (15) | 50.7% (38) | 22.6% (17) |
| Destroyed | 5.7% (7) | 4.9% (6) | 26.0% (32) | 63.4% (78) |

***Figure 3*. *Row-normalised confusion matrices for the MLP, Transformer, GP, QGP, and HQML-GP classification heads on the seen-event and unseen-event test sets*.** *Rows indicate the true damage classes and columns indicate the predicted classes. Cell values report the percentage*

*and number of samples assigned to each class, with each row normalised to 100%. Diagonal cells therefore represent class-specific recall, while off-diagonal cells show the distribution of misclassifications among no damage, minor damage, major damage, and destroyed.*

The unseen-disaster confusion matrices reveal a substantial change in class-specific behaviour. Recall decreases for the no-damage, major-damage, and destroyed classes across the evaluated heads, whereas minor-damage recall increases. The increase in minor-damage recall does not produce a corresponding improvement in minor-damage F1 because a larger number of no-damage buildings are also predicted as minor damage, reducing precision. The most pronounced shift therefore occurs along the boundary between no damage and minor damage.

For example, the MLP head correctly classified 84.5% of no-damage buildings on the seen-disaster test set but only 56.7% on the unseen event, with approximately one-third of unseen no-damage samples classified as minor damage. The QGP head retained the highest unseen-event no-damage recall at 68.3%, followed by HQML-GP at 65.2%. QGP also achieved the highest destroyed-class recall at 69.9%. Major-damage recall remained below 60% for every architecture on the unseen event, confirming that this class was the least stable under cross-event transfer.

Overall, Table 6 and Figure 3 identify two consistent patterns. First, no-damage and destroyed buildings are recognised more reliably than the intermediate minor- and major-damage categories. Second, transfer to the unseen Tuscaloosa event substantially increases confusion between adjacent severity levels, particularly between no damage and minor damage and between major damage and destruction. QGP achieved the strongest unseen-event recall for the no-damage and destroyed classes, but the GP-based heads did not consistently outperform the conventional baselines across every class. The principal unresolved challenge is therefore reliable discrimination of transitional damage states under event-to-event variation.

### 4.3. Model-specific qualitative assessment

Figure 4 presents representative cases in which exactly one of the five classification heads correctly predicts the ground-truth damage class, while the remaining four heads agree on the same incorrect class. The rows correspond, from top to bottom, to cases uniquely resolved by the MLP, Transformer, GP, QGP, and HQML-GP heads, respectively. The columns correspond to the four damage classes: no damage, minor damage, major damage, and destroyed. Unlike Figure 2, which illustrates samples on which all five heads either succeed or fail, Figure 4 isolates cases in which the choice of classification head changes the final prediction.

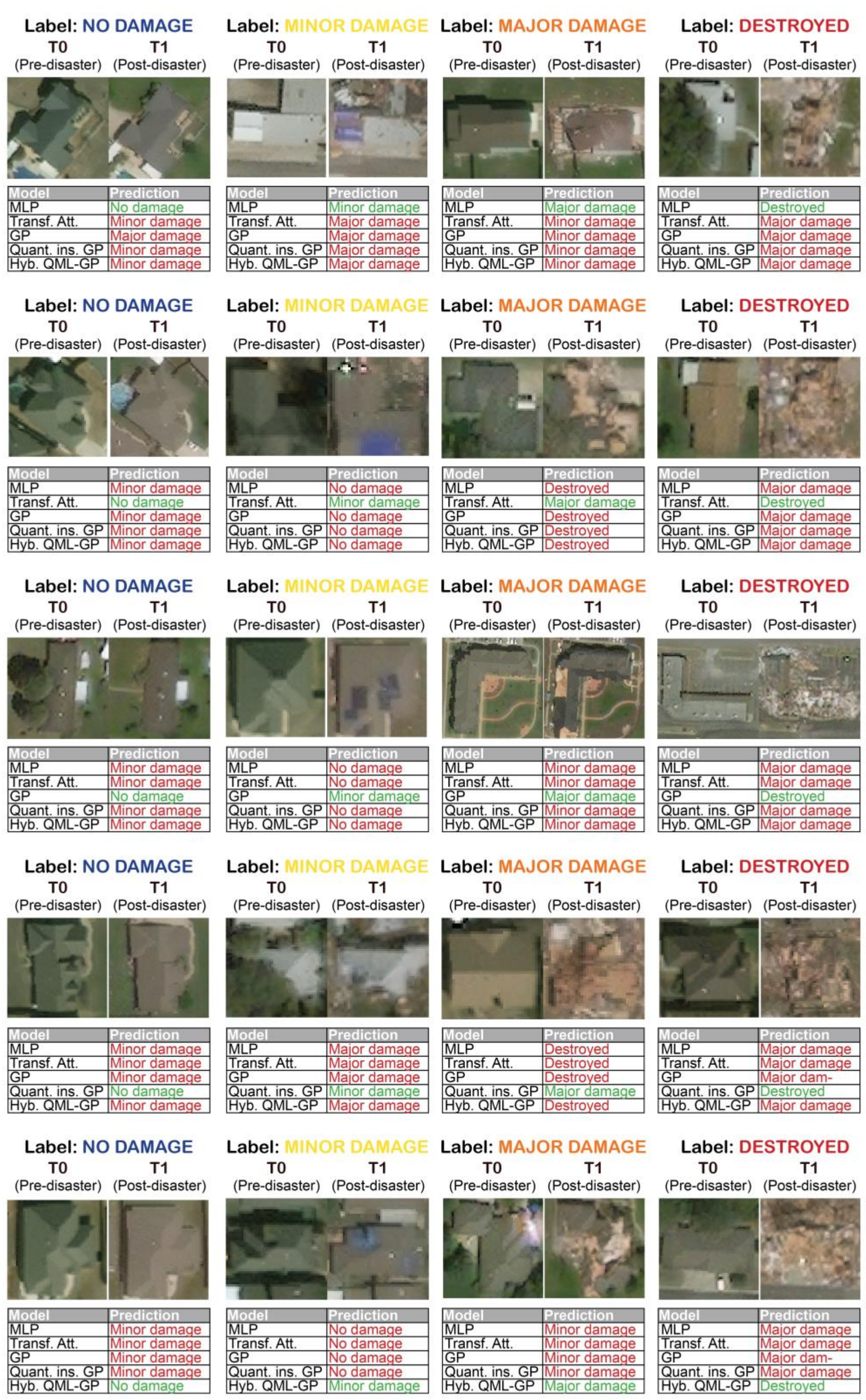

***Figure 4*. *Representative uniquely correct predictions produced by the five classification heads*.** *From top to bottom, the rows correspond to cases uniquely classified correctly by the MLP,*

*Transformer, GP, QGP, and HQML-GP heads, respectively, while the other four heads misclassified the same sample. From left to right, the columns show examples from the no-damage, minor-damage, major-damage, and destroyed classes. Green entries indicate correct predictions and red entries indicate incorrect predictions. The examples illustrate model-specific decision differences over the shared frozen ViT representation and should be interpreted qualitatively rather than as evidence of systematic superiority.*

Figure 4 shows that each classification head correctly resolves some samples that are misclassified by the other four heads. This pattern is observed not only for the proposed GP-based heads but also for the conventional MLP and Transformer baselines. The examples therefore do not indicate that one architecture is uniformly superior. Instead, they demonstrate that different head-level aggregation mechanisms can produce different decisions from the same frozen ViT token representation.

Several uniquely correct predictions occur in samples with visually subtle differences between neighbouring damage levels. This is particularly evident for the minor- and major-damage examples, where localised roof alteration, partial debris accumulation, surrounding destruction, or incomplete structural loss can support more than one plausible severity interpretation. In these cases, one head identifies the ground-truth class while the remaining heads jointly overestimate or underestimate the damage level. This qualitative pattern is consistent with the class-specific results in Table 6 and the confusion matrices in Figure 3, which identify adjacent-category confusion as the principal source of error.

Model-specific successes are also observed for the no-damage and destroyed classes. These examples suggest that differences among the heads may affect how evidence from the target building and its surrounding context is aggregated. However, the figure alone cannot establish which visual features caused a particular prediction or whether one head extracted additional information unavailable to the others. These examples therefore highlight differences in model behaviour under visually ambiguous conditions, rather than providing direct evidence of a specific internal representational mechanism.

Overall, Figure 4 indicates that no classification head is correct for every difficult case. The MLP, Transformer, GP, QGP, and HQML-GP heads each produce uniquely correct predictions for selected examples, suggesting that they form partially different decision boundaries over the shared frozen representation.

## 5. Discussion

The standard GP head produced competitive results relative to the MLP and Transformer baselines but did not consistently outperform them. On the seen-disaster test set, GP slightly exceeded the MLP in accuracy, weighted F1, and MCC and achieved a balanced accuracy close to the best-performing value. However, its unseen-event macro-F1 was lower than that of the Transformer and QGP heads. These findings indicate that explicitly representing token pair relationships through Plücker coordinates provides a viable alternative to mean pooling and self-attention, but the present evidence does not establish standard GP token mixing as uniformly superior to conventional aggregation mechanisms. Its primary contribution is therefore the introduction of a geometry-aware head-level representation that remains competitive under a controlled frozen-backbone comparison.

Among the evaluated architectures, QGP achieved the strongest point estimates on the overall test sets. On the seen-disaster test set, it obtained the highest accuracy, macro precision, macro-F1, weighted F1, and MCC. It also achieved the highest accuracy, macro-F1, weighted F1, and MCC on the held-out Tuscaloosa event. Relative to standard GP, QGP increased macro-F1 by 0.99 percentage points on the seen-disaster set and 2.25 percentage points on the unseen-disaster set. These results suggest that enriching normalised Plücker coordinates with amplitude-derived probability features may support more effective use of the geometric token representation under the reported experimental configuration. Furthermore, a component-level ablation would help quantify the individual contributions of amplitude normalisation, probability feature construction, and learnable feature scaling.

The class-specific findings provide a more differentiated interpretation of QGP performance. QGP achieved the highest no-damage F1 score in both evaluation settings and the highest minor-damage F1 on the seen-disaster test set. On the unseen event, HQML-GP achieved the highest minor-damage F1, although its advantage over QGP was only 0.2 percentage points. The Transformer head achieved the highest major-damage F1 on both test sets and the highest destroyed-class F1 on the unseen event. It also achieved the highest unseen-event balanced accuracy and macro precision. No single classification head therefore dominated every metric or damage category. QGP provided the strongest overall point estimates, while conventional self-attention remained particularly competitive for major-damage recognition and average class-level recall.

The behaviour of HQML-GP further clarifies the contribution of the simulated quantum circuit component. HQML-GP achieved the highest validation macro-F1, exceeding QGP by 0.59 percentage points, but this advantage did not transfer to either test set. It did not surpass QGP in overall seen- or unseen-event performance and required substantially greater training time because the differentiable quantum circuit was evaluated using a classical simulator. Under the present configuration, the addition of quantum circuit derived expectation values therefore provided no clear benefit on the test sets beyond the fully classical QGP representation. This finding does not rule out the potential value of alternative quantum encodings, circuit structures, or integration strategies, but it shows that greater architectural complexity did not automatically yield better predictive performance. A more pronounced pattern emerged across the damage categories than across the model architectures. All heads classified no-damage and destroyed buildings more successfully than minor- and major-damage buildings. The confusion matrices showed that errors occurred primarily between neighbouring severity levels. No-damage buildings were commonly confused with minor damage, while major-damage buildings were frequently confused with destroyed structures. Direct confusion between intact and destroyed buildings was comparatively uncommon. These results reflect the visual and semantic ambiguity of intermediate damage states, where localised roof alteration, partial structural loss, debris, shadow, surrounding destruction, and image-acquisition conditions can support more than one plausible class interpretation. Major damage was the least stable class under transfer to Tuscaloosa. Although the standard GP head achieved the highest seen-event major-damage recall, the Transformer achieved the highest major-damage F1, indicating a more favourable balance between precision and recall. On the unseen event, major-damage F1 remained low for every architecture. This suggests that the central unresolved problem is not merely the selection of a stronger classification head, but the reliable representation and labelling of transitional damage states under event-to-event variation. Given the ordinal structure of the four damage categories, standard cross-entropy treats all misclassifications equally and therefore does not distinguish between adjacent-class errors and more severe errors spanning multiple damage levels. Confusing major damage with destruction is not equivalent to confusing major damage with no damage, yet both errors receive the same penalty in the current objective.

All models experienced substantial performance degradation on the held-out Tuscaloosa tornado. QGP retained the highest overall accuracy, macro-F1, weighted F1, and MCC, but its unseen-event

macro-F1 exceeded that of the Transformer by only 0.20 percentage points. Furthermore, the Transformer achieved higher balanced accuracy and macro precision on the same event. The results therefore support the conclusion that QGP was competitive and achieved the strongest overall point estimates on Tuscaloosa, rather than establishing general cross-disaster robustness. Evaluation on a single held-out tornado cannot determine whether the observed ranking would remain stable across other tornadoes, geographic regions, hazard types, building typologies, or image-acquisition conditions. The qualitative analyses reinforce this interpretation. Figure 2 shows that all heads succeed when damage evidence is visually clear and can fail together when the target building is affected by ambiguous local or contextual features. Figure 4 further demonstrates that each architecture can correctly resolve selected cases misclassified by the other heads. These examples indicate that the MLP, Transformer, GP, QGP, and HQML-GP heads form partially different decision boundaries over the same frozen representation. However, the selected images are illustrative examples and do not establish how frequently each architecture produces uniquely correct predictions. The quantitative results in Tables 5 and 6 therefore remain the primary evidence for comparing the evaluated heads. From a computational perspective, head-level adaptation avoids updating the full ViT backbone and limits trainable parameters to the classification component. Within this protocol, the MLP remained the smallest head, while the GP-based heads required fewer than 0.8 million trainable parameters. QGP achieved the most favourable observed balance between overall predictive performance and the reported computational cost, whereas HQML-GP incurred substantially longer training time without improving test performance over QGP. The reported inference-time differences should nevertheless be interpreted cautiously because they are implementation- and hardware-dependent, and mixed precision was disabled for HQML-GP but enabled for the classical heads.

## 6. Conclusions

This study introduced three geometry-aware classification heads—Grassmann–Plücker token mixing (GP), Quantum-inspired Grassmann–Plücker token mixing (QGP), and Hybrid Quantum Machine Learning Grassmann–Plücker token mixing (HQML-GP)—for four-class building damage assessment from paired pre- and post-disaster satellite imagery. All heads were evaluated using the common frozen six-channel ViT-B/16 encoder applied to paired pre- and post-disaster xBD tornado imagery, under identical data partitions, optimisation procedure, checkpoint selection criterion, and evaluation protocol. This controlled design isolated differences among the head architectures while holding the frozen visual representation constant.

Quantum-inspired enrichment produced a further and consistent gain. QGP led five of six aggregate metrics on the seen-event test set and four of six on the held-out Tuscaloosa tornado. Its advantage was concentrated where it matters operationally: the highest recall on both no-damage and destroyed buildings under event transfer. The geometric properties of the construction are relevant to this behaviour. To our knowledge this is the first demonstration on a real remote sensing benchmark that a quantum-inspired token mixing operator outperforms a self-attention module of comparable size under an identical frozen representation and a fully matched training protocol.

HQML-GP achieved the highest validation macro-F1, confirming that differentiable quantum circuit features can be integrated successfully with Grassmann–Plücker token mixing. However, this validation advantage did not translate into superior test performance relative to QGP, while the simulated quantum circuit incurred substantially greater training cost. The findings therefore do not support additional hybrid quantum–classical complexity as automatically beneficial. Instead, they position HQML-GP as a promising exploratory extension and identify the fully classical QGP formulation as the more effective configuration under the present experimental conditions.

More broadly, the results provide a cautious but practically relevant perspective on the role of quantum and quantum-inspired learning within contemporary computer vision systems. Quantum and hybrid quantum–classical models have not yet demonstrated consistent superiority over strong classical deep learning methods for large-scale classical data, owing to challenges associated with data encoding, circuit trainability, scalability, hardware noise, and the computational cost of quantum simulation [74-77]. Rather than attempting to replace an established vision transformer architecture, this study introduced quantum-inspired and hybrid quantum–classical operations selectively within the classification head. Under this controlled protocol, QGP outperformed both standard GP and the conventional Transformer head while maintaining comparable inference efficiency. HQML-GP further demonstrated the feasibility of injecting quantum circuit expectation values into geometric token representations, although its additional complexity did not produce stronger test generalisation. These findings support lightweight and targeted quantum-inspired enrichment of strong classical representations as a more immediately practical direction than end-to-end replacement of mature deep learning architectures.

The results also reveal important unresolved challenges. Performance declined substantially on the held-out disaster, and all models continued to experience difficulty distinguishing visually ambiguous intermediate damage states. The findings are further bounded by the use of one fixed split, a single random seed, and one unseen tornado event. Future research should assess run-to-run variability across multiple seeds, extend evaluation through multi-event and leave-one-event-out experiments, and conduct component-level ablations to quantify the contributions of GP offsets, amplitude normalisation, probability feature construction, feature scaling, and quantum circuit configuration. Ordinal and imbalance-aware objectives, parameter-efficient backbone adaptation, predictive uncertainty, calibration, and comparisons between hybrid circuits and classical modules of comparable complexity also warrant investigation.

## DATA STATEMENT

All data used in this study are publicly available through the xBD dataset.

## CRediT author statement

**K. Farahkhah, U. Lagap:** Methodology, Visualization, Validation, Formal analysis, Data curation, Writing-Original Draft; **T. Rezaei:** Visualization, Writing-Original Draft; Writing-Reviewing and Editing, Supervision; **S. Ghaffarian:** Conceptualization, Methodology, Writing-Reviewing and Editing, Supervision, Project administration

## DECLARATION OF COMPETING INTERESTS

The authors declare that they have no known competing financial interests or personal relationships that could have influenced the work reported in this paper.

## DECLARATION OF GENERATIVE AI USE

The authors used ChatGPT by OpenAI solely for language editing and proofreading. The authors reviewed and edited the output and take full responsibility for the content of this publication.